\documentclass{article}[15pt]

\usepackage{amsmath,epsfig,color,calc,rotating,graphics,ulem}

\usepackage{url}

\usepackage[pagewise]{lineno}

\usepackage{natbib}
\bibpunct{(}{)}{;}{a}{}{,}

\usepackage{url}
\usepackage{float}

\begin{document}

\title{
Non-Zipfian Distribution of Stopwords or Function Words and Subset Selection Models
\vspace{0.2in}
\author{
Wentian Li$^{1,2}$\footnote{Corresponding author: wentian.li@stonybrook.edu} and Oscar Fontanelli$^3$  \\
{\small \sl 1. Department of Applied Mathematics and Statistics} \\
{\small Stony Brook University, Stony Brook, NY, USA } \\ 
{\small \sl  2. The Robert S. Boas Center for Genomics and Human Genetics}\\
{\small \sl  The Feinstein Institutes for Medical Research, Northwell Healths,  Manhasset, NY, USA}\\
{\small \sl  3. Methods Lab, Facultad Latinoamericana de Ciencias Sociales,  M\'{e}xico }\\
}
\date{\today}
}
\maketitle  
\markboth{\sl Li and Fontanelli}{\sl Li and Fontanelli}

\newpage


\Large

\begin{center}
Abstract
\end{center}

\normalsize

Stopwords and function words are relatively less informative for the content of 
a language and more often play a structural role in a sentence. 
Stopwords are ubiquitous words and may contain verbs, adjectives and adverbs.
On the other hand, function words are strictly
prepositions, conjunctions, pronouns, determiners, qualifiers, articles, 
interrogatives, and a limited number of auxiliary  verbs.
In contrast to the well known Zipf's law for rank-frequency plot for all words, 
the rank-frequency plots for stopwords or function words 
are best fitted by the Beta Rank Function (BRF).
On the other hand, the rank-frequency plots of non-stopwords or non-function-words 
also deviate from the Zipf's law, but are better described by a quadratic
function of log-token-count over log-rank than by BRF. Based on the observed rank 
of stopwords or function words in the full word list, we propose
a stopword/function word/subset selection model that the probability 
for being selected, as a function of the word's rank $r$, is a decreasing Hill's
function ($1/(1+(r/r_{mid})^\gamma)$);
whereas the probability for not being selected is the standard Hill's
function ( $1/(1+(r_{mid}/r)^\gamma)$).
We validate this selection probability model by a direct estimation
from an independent collection of texts.
We also show analytically that this model leads to a BRF rank-frequency
distribution for stopwords or function words when the original 
full word list follows the Zipf's law, as well as explaining the quadratic fitting 
function for the non-stopwords or non-function-words.
A corollary of these results is that Zipf's law is
not expected to be true for telegraphic speech in early childhood 
language learners or in agrammatism patients.

\vspace{0.2in}
 
 
\vspace{0.2in}

{\bf Abbreviations:} 
BRF: Beta Rank Function;
NLP: natural language processing;
NLTK: Natural Language Toolkit;
NSFW: non-stopwords and/or non-function-words;
QL: quantitative linguistics;
SFW: stopwords and/or function words;
 
\large

\newpage

\section{Introduction}

\indent

Stopword \citep{wilbur,gerlach} is a concept in natural language processing 
(NLP) referring to those words that might be filtered out before a text analysis task
because they carry very little semantic meaning.  An operational
definition of stopwords is that these are ubiquitous. These may appear in each
chapter in a book, in most documents in a corpus, many paragraphs in an article, etc. 
The ubiquity can be measured by a zero or low level of the inverse document frequency 
\citep{robertson}, which is a measure of uniqueness, the opposite of ubiquity.
These objective definitions can extend the concept of stopwords to even technical
or mathematical texts \citep{sarica}. 

Function words \citep{fries} or structural words or grammatical words 
are proposed in linguistics as one class of words that are different from the content words
\citep{winter53}.
The set of function words do not evolve in time (close class). They play a role
of grammatic ``glue" to other words. Only these types of words belong to 
the function words class:
prepositions (in, of, with, for),
conjunctions (and, or, but),
pronouns (I, he, me, other), 
determiners including articles and demonstratives and numbers (a/an, the, this, my, some, one, first),
qualifiers (very, more, quite),
and auxiliary verb (can, will, have, be, do).

Stopwords and function words are an antithesis of the index words
\citep{mulvany} often found in the last few pages of a book, or keywords
often listed on the first page of a scientific paper \citep{shah}.
Listing stopwords or function as index words or keywords would make 
no sense as they most likely do not provide unique information.
Of course, the distinction between function words and
content words, between stopwords and non-stopwords, can always be under 
debate for individual cases \citep{semi}. For convenience, we use SFW to
abbreviate ``stopwords and function words" and NSFW for ``non-stopwords
and non-function-words" in this article.

While SFWs are  relatively less useful 
for NLP, ironically, these are very important
in quantitative linguistics (QL) (as well as for authorship/stylometry analysis
\citep{kestemont,segarra,neal}). When dictionary words are ranked according
to their number of appearance in a text from high to low, SFWs dominate
the highest-ranking list. Roughly 80\%$\sim$90\% of top-100 most common
words are SFWs, depending on which SFW list and which source text
is used (see the Results section).  
As it is well known that the number of appearance (number of tokens per type)
as a function of rank (rank-frequency or rank-size plot) is roughly an inverse 
power-law function with exponent
close to 1, or Zipf's law \citep{zipf35,wli-every,piantadosi}, the existence, or absence for that
matter,  of stopwords should impact the ``head" part of the rank-frequency plot.

When SFWs are ranked within the group, one may redraw the
rank-frequency plot for SFWs only. What would be the functional
form for the SFW-only rank-frequency plot? Is it still a Zipf's law?
In this paper, we address this question both through text analyses
and analytic solutions. It is noticed that the question posed is essentially 
about a subset when the whole dataset is already known to follow a 
specific (e.g. Zipf's) distribution. Depending on how
the subset is sampled, generally speaking the subset may not follow
the same rank-frequency distribution as the original dataset where
they are sampled from \citep{cristelli}.

A general argument concerning the rank-frequency plot of a subset from
a power-law distributed dataset is the following. Suppose the full dataset
follows exactly Zipf's law:
\begin{equation}
T(r) = \frac{c}{r^\alpha}
\end{equation}
where $T$ is the number of tokens per dictionary word type, $r$ is the rank,
and $\alpha \approx 1$ is the exponent. When a subset of samples are
selected, they have new ranks, $r_{new}$, within the subset itself. If there
is a smooth function that maps $r_{new}$ to the original rank $r$: $r=g(r_{new})$,
then the rank-frequency plot for samples in the subset is:
\begin{equation}
\label{eq-r2rnew}
T(r_{new}) = \frac{c}{(g(r_{new}))^\alpha}
\end{equation}
which can still be a power-law Zipf function if $r$ is a power-law 
function of $r_{new}$, or equivalently, $r_{new}$ is a power-law 
function of $r$ (linear function included), but will  no longer be a Zipf's law
otherwise. 

In this work, we empirically obtain a functional form between
the SFW  ranks within the subset and in the original full set,
and use that knowledge to address the issue of rank-frequency plot of
SFWs.

\section{Data and Methods}

\subsection{List of stopwords and function words}

\indent 

There is no universally agreed stopword list, partly because
whether a word is ubiquitous or not depends on the context
and documents used. In the case of function word list,
whether rare, e.g., prepositions should be included is uncertain. In an extreme example,
since all numbers are part of the function words, to include all numerical names would
lead to infinite number of function words in the list. We rely on public-domain
stopword lists as the basis, then a function words list is constructed by combining
multiple lists followed by removing verbs, nouns, adjectives, etc.  
In a compilation of over 60 stopwords lists
(\url{https://github.com/igorbrigadir/stopwords} )
the number of stopwords can be as low as a little bit more than 20,
to over 1000.

A common understanding of stopwords is that if a word is ignored in a search 
and indexing, that word would be a stopword for this specific search task.
Examples include MySql database search 
\url{https://dev.mysql.com/doc/refman/8.4/en/fulltext-stopwords.html}
and early day's web search engines. However, some recent search engines, 
such as those based on bidirectional encoder representations from transformer (BERT),
are able to extract subtle information from all words and do not need
to distinguish stopwords from non-stopwords \citep{bert}

We chose to use the following two stopwords lists.
The first is from NLTK (Natural Language Toolkit) \citep{loper},
\url{ https://www.nltk.org/nltk_data/}, for
``stopwords corpus", URL:
\url{https://raw.githubusercontent.com/nltk/nltk_data/gh-pages/packages/corpora/stopwords.zip}. 
The file ``english" has 198 entries. Among them, 74  are either
contracted forms (e.g., i've) or part of contracted
forms (e.g., ``ain" and ``t" as part of ain't), leaving
124 words in non-contrasted forms.

The second list is from an open-source library  spaCy
\url{https://spacy.io/} \citep{vasillev}, URL:
\url{https://github.com/explosion/spaCy/blob/master/spacy/lang/en/stop_words.py} .
There are 305 stopwords in the list and none of them are in  contracted forms.
From the list, we removed the entry ``re", ``ca", ``thru",
with 302 stopwords left.
Of 124 non-contrasted NLTK stopwords, 122 are in the spaCy list,
whereas 180 spaCy stopwords are not in the NLTK list. 

A third stopword list is used to validate our subset selection model
but not used in the initial pattern-finding analysis: the snowball 
(\url{https://snowballstem.org/}) stopword list. The English language
component of that list contains 175 stopwords, available from the {\sl R} package
{\sl stopwords} (\url{https://cran.r-project.org/web/packages/stopwords/}).

As for function words list, we combined the NLTK and spaCy 
stopword lists plus another list with about 200 common words
(Appendix B of \citep{craig}, or  \\
\url{https://doi.org/10.1017/CBO9780511605437.013}).
Nouns, verbs (except for auxiliary verbs),
adjectives, adverbs (except for those with a structural role), and numbers larger
than eight (except ``ten"), are then removed to form a function words list
with 228 entries.

\subsection{Text files}

\indent

The Brown Corpus \citep{brown} was downloaded through NLTK
\url{https://www.nltk.org/nltk_data/}, at
\url{https://raw.githubusercontent.com/nltk/nltk_data/gh-pages/packages/corpora/brown.zip},
with more than 1.1 millions tokens. 
The number of word types is 47437.

The {\sl Moby Dick} text was downloaded from the
Project Gutenberg
\url{https://www.gutenberg.org/ebooks/2701}, with more than 210,000 tokens.
The number of word types is around 20000.

For validation of our subset selection model, we used 30 books from
the Project Gutenberg. These 30 books are:
1*. Alice in Wonderland;
2. Anne of the Green Marbles;
3*. Beowulf;
4*. Bleak House;
5. Confessions (St. Augustine);
6. Crime and Punishment;
7*. Discourse on the Method;
8*. Don Quixote;
9*. Dracula;
10*. Essay on Human Understanding;
11*. Frankenstein;
12*. The Great Gatsby;
13*. Hamlet;
14. Heart of Darkness;
15. Huckleberry Finn;
16. The Iliad;
17*. Inquiry on the Wealth of Nations;
18*. Leviathan;
19.  Metamorphosis (Kafka);
20*. Middlemarch;
21*. Moby Dick;
22*. Natural History (Plinius);
23*. Origin of the Species;
24.  Portrait of Dorian Gray;
25*. Pride and Prejudice;
26. The Prince;
27*. Ulysses;
28*. Utopia;
29*. War and Peace;
30. Wuthering Heights.
The 20 titles with asterisk are also used in \citep{wli-heaps} and were described
with more detail there.

\subsection{Fitting rank-frequency plots by regressions}

\indent

When the number of tokens ($T$) per word is ranked from high to low,
rank ($r$) is 1 for the most frequently appearing word, $r=2$ is the
second most frequent word, etc. The $T$  ($y$-axis) vs. $r$ ($x$-axis)
is the rank-frequency plot, often in log-log scale.

The following fitting functions are used: (1) Zipf's law which is $T= c/r^\alpha$ or
\begin{equation}
\label{eq-zipf}
\log(T)= c' - \alpha \log(r) 
\end{equation}
in log-log scale; (2) quadratic correction of the power law:
\begin{equation}
\label{eq-quadratic}
\log(T)= c' - \alpha \log(r) - \kappa (\log(r))^2
\end{equation}
(3) discrete generalized beta distribution (DGBD)
\citep{mansilla,martinez,alvarez,oscar-city,nowak}
or  beta rank function (BRF) 
\citep{oscar-brf,wli-C,oscar-lavalette,wli16,oscar-covid}
(also called beta-like function \citep{naumis},
Cocho rank function \citep{wli-letter},
beta function \citep{wli-entropy}, etc):
$T= c(r_{max}+1-r)^\beta/r^\alpha$, or, in log-log scale, 
\begin{equation}
\label{eq-brf}
\log(T)= c' - \alpha \log(r) +\beta \log(r_{max}+1-r)  
\end{equation}
This function appeared in \citep{gilchrist} as proposed
by Davis, in \citep{hankin} by Hankin and Lee,
and independently discovered by Prof. Germinal Cocho and his colleagues
in \citep{mansilla}.
(4) The Mandelbrot function or generalized Zipf's law \citep{mandelbrot}: 
\begin{equation}
\label{eq-mand}
\log(T)= c'- \alpha \log(r+B).
\end{equation}

\subsection{Evenly sampled data points in log-scale}
\label{sec-log}

\indent

In order to fit a rank-frequency plot by these two functions, one
tricky point, rarely discussed in the literature, is that in order
to visually see the fitting function overlapping the data points,
it might be necessary to not use all data points in the plot,
but only some of them.  The reason is that data points
in a rank-frequency plot in log-log scale are not evenly distributed
in the $x$-axis. Using all data points may favor the tail area
(rare words with large $r$ values) of the rank-frequency plot,
sacrificing the head area (common words with high rankings). 

To solve this problem, we can use evenly sampled data points
in the log scale. For example, picking data points with ranks
(using a $R$ code) \texttt{c(1:10, unique(round(10*1.05\textasciicircum{}c(1:155)))) }
will span the range of rank from 1 to $10 \times 1.05^{155} \sim 19246 $,
but sample them evenly in the log scale, and at the same time keep the
top ten words. The \texttt{round} function will convert real values
to integers, and \texttt{unique} function will combine same rank
values due to the conservative geometric expansion in $r$ 
(controlled by the value 1.05 in the above example).  

\subsection{Brute force nonlinear function fittings}
\label{sec-grid}

\indent

For nonlinear fitting that can not be converted to a linear fitting
with unknown parameters as coefficients, we use a ``grid" method
or brute force method.
In this method, we fix one parameter, running a linear fitting to
get other parameters fitting values, record the fitting performance,
then change the parameter and rerun the fitting, repeatedly;
or, we fix one parameter, running through a limited
number of values for the second parameter, find the best choice,
then repeat the process for another setting for the first parameter. 
For example, in Eq.(\ref{eq-mand}) we fix the $B$ value before 
running a function fitting.

\subsection{Programs used}

\indent

All scripts were written by the authors in the $R$ statistical platform
\url{https://www.r-project.org/}.
All linear function fittings were carried out by the {\sl lm()}
(linear model) in {\sl R}. 
For the validation of our subset selection model, we used
the nonlinear least square function ({\sl nls} in {\sl R}) using
the ``port" algorithm. 

\subsection{Sample R-scripts for distribution}

\indent

For the purpose of reproducibility, R-scripts,
as well as the required data, for some major calculation in
this paper can be found at
\url{https://github.com/wlicol/stopwords}.

\section{Results}

\subsection{ Relationship between different SFW lists}

\indent

There are 122 stopwords shared between NLTK and spaCy lists. Among these,
there are 28 pronouns (e.g., he, it, I), 25 prepositions (e.g., if, to, in), 
18 auxiliary verbs (e.g., is , can, has,  do), 14 conjunctions (e.g., and, as, but),
13 quantifiers (e.g., more, only, very),
8 demonstratives or ``W" words (e.g., which, when, who), 6 determiners (e.g., this,
some, any), 6 adverbs (e.g., now, there, again), 3 articles (a, the, an),  and
1 adjective (own). Some words may belong to different
classes, thus some parts of the above partitions are debatable.

The NLTK stopwords that are not in spaCy list 76 of them) are all contracted words
(e.g., aren't) or their components (e.g. aren, t). On the other hand,
those spaCy stopwords not in the NLTK list (180 of them) expand to include
more adverb (e.g., even, also, well, never), 
auxiliary verbs (e.g. would, could, may), other verbs (e.g., give, make, see, get, go, take, move),
prepositions (e.g. without, around, per, within),
numerals (e.g., one, two, first, ten, hundred), 
determiners or qualifiers (e.g., many, much, enough, next),
pronouns (e.g. us, anything, everyone, none, nothing),
conjunctions (e.g. although, however, since, whether, either, unless),
nouns (part, side, name, amount, top, bottom),
``W" words (e.g., whose, whatever, whoever, whereby),
adjectives (e.g. every, whole, full, last, various, serious),
and articles (every).

In order to construct a function word list,
we first removed 84 words from the spaCy list, with 218 words left.
Then, 8 function words from \citep{craig} (like, past, round, shall,
somewhat, till, unto, whilst) and 2 function words from NLTK
(having, theirs) were added, resulting in a list of 228 function words.

We used another stopword list, the snowball stopwords,
in the validation of our subset selection model, to be discussed in
later subsections. Besides the 50 contracted words which are not included
in NLTK-spaCy common and in spaCy, these three snowball stopwords
are not in the common and spaCy list: having, ought, theirs.
For convenience, the 122-stopword (common stopwords between NLTK and spaCy),
the 302-stopword (spaCy) lists, 228 function words,
the 175-stopwords (snowball), are included in the Appendix.

\subsection{The rank-frequency plot of stopwords follows a beta rank function}

\indent

We used two texts (a single novel {\sl Moby Dick} and a whole Brown corpus)
and  three SFW lists (122 common stopwords 
between NLTK and spaCy, 302 stopwords in spaCy, and 228 function words) 
to examine the rank-frequency distribution of SFWs. 
SFWs tend to rank higher than NSFWs.
46 (92\%) of the top 50 words in {\sl Moby Dick}
and 48 (96\%) of the top 50 words in Brown corpus belong
to the 122 common stopwords list. These percentages are only slightly
reduced to 75\% and 78\% for the top 100 words in two text sources.
For the 302 spaCy stopwords list, these four percentages become
96\% (top 50 words in {\sl Moby Dick}), 100\% (top 50 words in Brown corpus), 
85\% (top 100 words in {\sl Moby Dick}), and 92\% (top 100 words in Brown corpus). 

\begin{figure}[H]
 \begin{center}
 \includegraphics[width=1.0\textwidth]{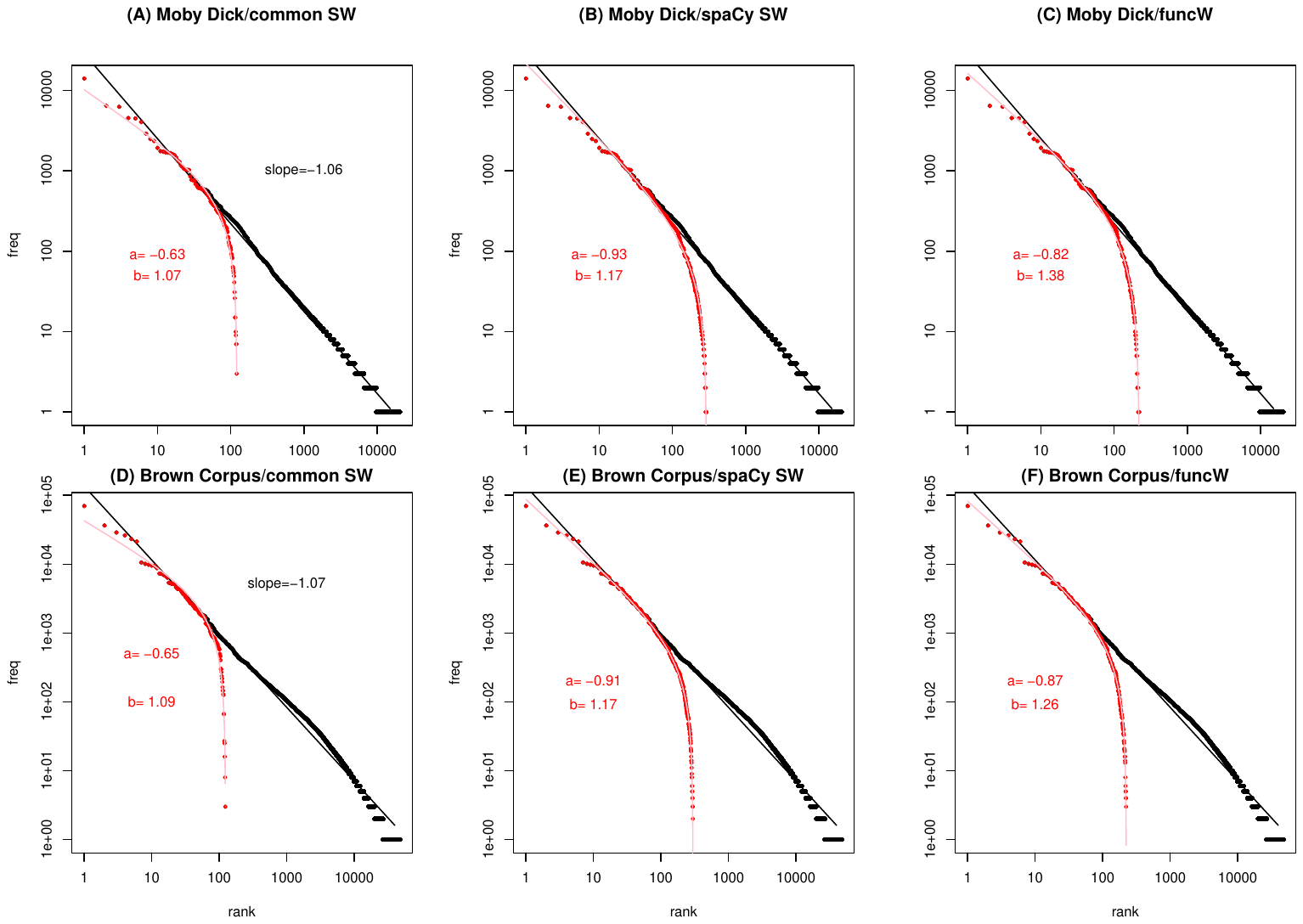}
 \end{center}
\caption{ \label{fig1}
Rank-frequency plot of all words (black) from {\sl Moby Dick}
(top row) and Brown corpus (bottom row), and stopwords (red)
for NLTK (left column, $n=$122 after removing the contracted words and/or their
components), stopwords for spaCy (middle column, $n=$302) lists,
and 228 function words (right column). The fitting 
lines for all words are power-law (Zipf) function, and those for
stopwords or function words (pink lines) are BRFs. 
Note that there are six plots
for stopwords or function words because the number of combinations 
between SFW lists and source texts is 6.
}
\end{figure}

Fig.\ref{fig1} shows the 6 combinations 
(three SFW lists and
two text sources)  of rank-frequency plot in log-log scale.
The rank-frequency of all words for {\sl Moby Dick} follows the
Zipf's law almost perfectly, with a fitting exponent of $\alpha=1.06$.
That for the Brown corpus is reasonably Zipf's (using the evenly
sampling method for log scale discussed in the Method section),
with $\alpha=1.07$, but a bump from rank ranging from a few hundreds
to ten thousands can be visually seen, indicating that Zipf's law
is not perfect.

For SFWs, their number of appearances are the same as before,
but their rank values are lower, because the new rank is for
stopwords only. This effectively moves these data points horizontally
to the left in the rank-frequency plot (see Fig.\ref{fig6}(A), to be
discussed in the Discussion section). The resulting line
(see Fig.\ref{fig1}) is curved and can be perfectly fit by the
beta rank function. The fitting values for $\alpha$ are 
0.63, 0.93, 0.82, 0.65, 0.91, and 0.87 for these 6 
combinations, and for $\beta$ are
1.07, 1.17, 1.38, 1.09, 1.17, and 1.26.

\subsection{A subset sampling model explaining the beta rank function in stopword
rank-frequency plot
\label{sec-ssm}}

\indent

We propose a model to explain the BRF in the rank-frequency
plot of stopwords, starting from an ideal Zipf's law, followed by a
subset selection.
Graphically, the initial Zipf's law rank-frequency plot is a
straight line with negative slope in the log(rank)-log(frequency)
scale. Converting the power-law to a BRF is to curve the line
which can be achieved 
by moving points leftwards (Fig.\ref{fig6}(A)). This approach
will reduce the number of points in the curve, which can only
be achieved by removing points, or keeping only a  subset of
points. 

\begin{figure}[H]
 \begin{center}
 \includegraphics[width=1.0\textwidth]{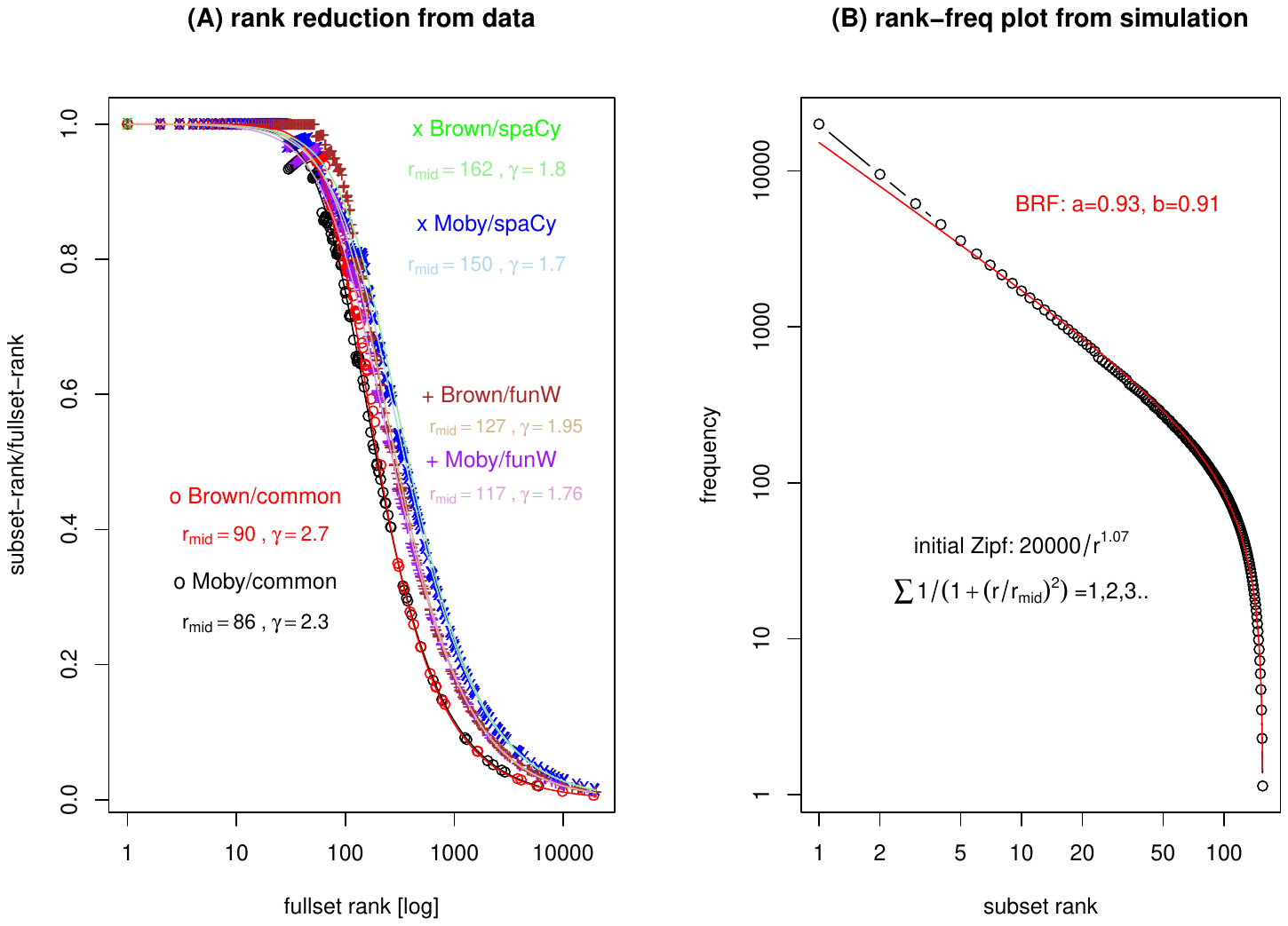}
 \end{center}
\caption{ \label{fig2}
Subset sampling model.  A rank-$r$ may be selected (subset selection)
with a new rank $r_{new}$ within the subset.
(A) $r_{new}/r$ as a function of $r$ in 4 combinations of 2 text sources
({\sl Moby Dick} and Brown corpus) and 3 stopword/function-word lists. 
Six cumulative sums of  the decreasing Hill function Eq.(\ref{eq-subset-model}))
are shown that best fit the six sets of data.
(B) Rank-frequency plot of a simulated dataset from the subset selection model.
The red line is a BRF fitting function.
}
\end{figure}

To get an idea on how keeping only the SFWs would affect their rank,
we plot the rank reduction ratio, $r_{new}/r$, where $r_{new}$
is the rank among SFWs only, and $r$ is the original rank
in the full set,  as a function of
the original rank $r$ (in log scale) in Fig.\ref{fig2}(A),
for the six source/SFW-list combinations in Fig.\ref{fig1}.

Based on the plot in Fig.\ref{fig2}(A), we consider
the change to be selected as a stopword has the form of 1 minus
logistic function with the log-$x$ as the independent variable.
More precisely, the probability of a rank-$r$ word to be a stopword
is modelled as:
\begin{equation}
\label{eq-subset-model}
Prob(stopword)_r= 1- \frac{1}{1+exp( - \gamma \log( \frac{r}{r_{mid}}) )}
= \frac{1}{1+exp(\gamma \log( \frac{r}{r_{mid}} ) )}
= \frac{1}{1+ \left(\frac{r}{r_{mid}} \right)^\gamma}
\end{equation}
The $r_{mid}$ is the rank value where the probability of being selected is 0.5.
The functional form in Eq.(\ref{eq-subset-model}) is called 
(decreasing or repressive form of) Hill's function/equation \citep{hill}
used in pharmacology and biochemistry.
The new rank $r_{new}$ in a rank-$r$ selected word is a cumulative
sum of the probabilities to be selected up to rank $r$, in this model:
\begin{equation}
\label{eq-cumsum1}
r_{new} = \sum_{u=1}^r \frac{1}{1+ ( \frac{u}{r_{mid}} )^\gamma}
\approx \int_{u=1}^{u=r} \frac{du}{1+ ( \frac{u}{r_{mid}} )^\gamma} .
\end{equation}

We have obtained the best fit parameter values for $(r_{mid}, \gamma)$ for 
Moby Dick/common stopwords to be (86, 2.3), 
(90, 2.7) for Brown corpus/common stopwords, 
(150, 1.7) for Moby Dick/spaCy stopwords, 
(162, 1.8) for Brown corpus/spaCy stopwords, 
(117, 1.76) for Moby Dick/function words, and
(127, 1.95) for Brown corpus/function words.
The corresponding curves by Eq.(\ref{eq-cumsum1})
are black, red, lightblue, lightgreen, plum, and tan in Fig.\ref{fig2}(A),
and they fit the data almost perfectly.

\subsection{Validation of stopwords selection probability model}

\begin{figure}[H]
 \begin{center}
 \includegraphics[width=1.0\textwidth]{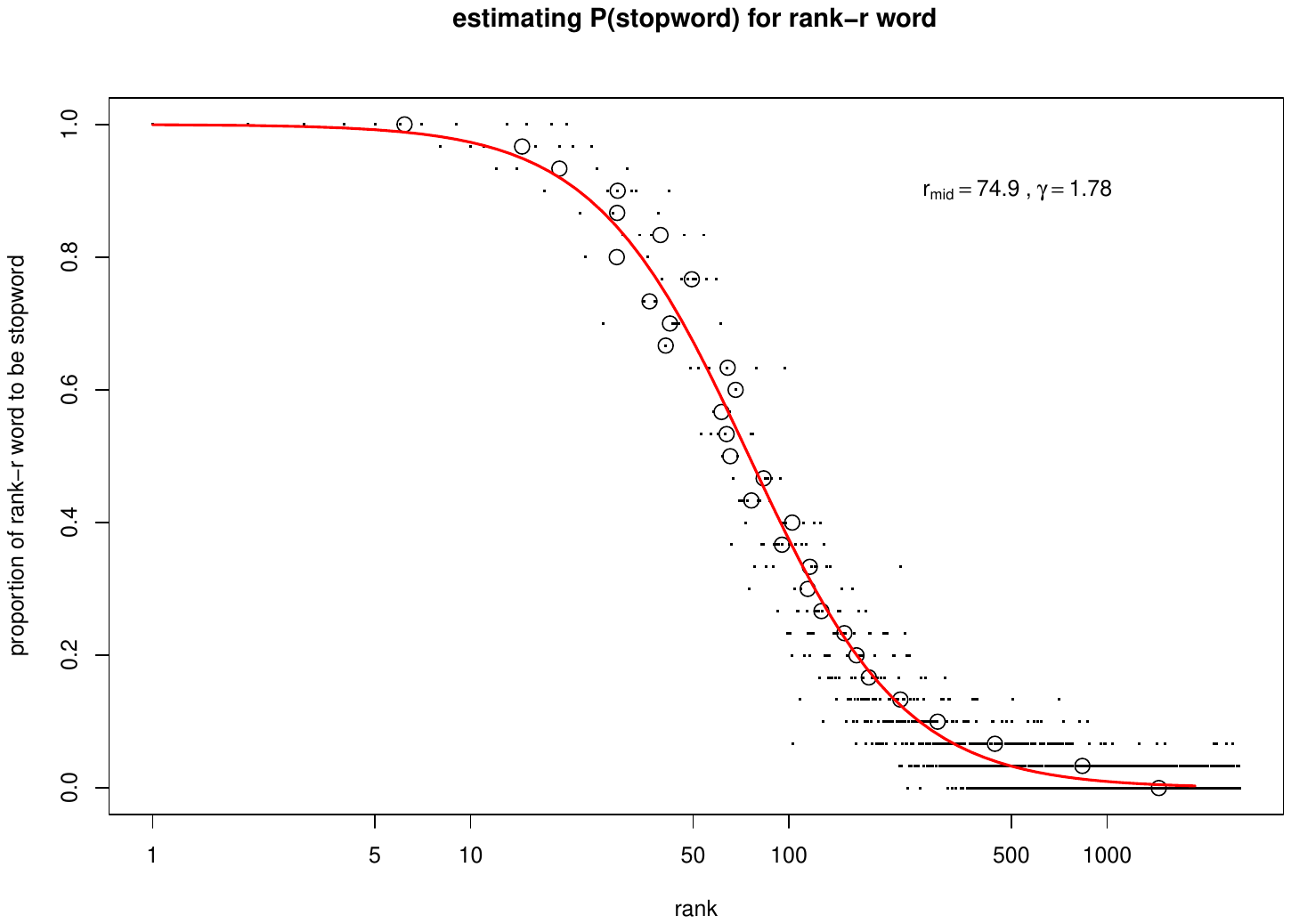}
 \end{center}
\caption{ \label{fig3}
We use 30 books to estimate the probability of a rank-$r$ word
to be a stopword. Each dot shows the statistics of rank-$r$ words:
the $y$ is the proportion of 30 rank-$r$ words that are stopwords,
and $x$ is the rank $r$. Circles represent the geometric mean of dots with
the same $y$ value. A nonlinear function fitting of Eq.(\ref{eq-subset-model})
leads to the estimation of parameters: $\gamma=1.78$ and $r_{mid}=74.9$.}
\end{figure}

\indent

Our formula in Eq.(\ref{eq-subset-model}) for a rank-$r$ word to be a SFW
is based on the progression of $r_{new}/r$ in one text source. Here we test
its validity by using 30 text sources (books or book-length texts from
Project Gutenberg -- 20 of them are used and described in \citep{wli-heaps}). 

For each source of text, words are ranked by their number of tokens in the
text. We then check if a rank-$r$ word is a (snowball) stopword or not.
The probability of rank-$r$ word to be a stopword is calculated directly 
from these 30 rank lists by the proportion of 30 rank-$r$ words,
that belong to the stopword list. Note that a (e.g.) rank-30 word
in one book may not be the same as the rank-30 word in another book,
and even if both are stopwords, they may or may not be the same word. 

Fig.\ref{fig3} shows the proportion of rank-$r$ words that are stopwords
in the 30 text sources as a function of the rank $r$ (small dots). To
smooth the scatter plot, we take the mean of log-rank for words with 
the same stopword proportion (in large circles). These are fitted
by Eq.(\ref{eq-subset-model}) using nonlinear function fitting. The estimated
parameter values are: $r_{mid}$ around 75, and $\gamma=1.78$. By using an independent
stopword list and an independent collection of 30 texts, we validate
our stopword selection probability model Eq.(\ref{eq-subset-model})
by a direct estimation.

\subsection{Analytic proof that the subset selection model leads to BRF
distribution}

\indent

With the subset selection model of Eq.(\ref{eq-subset-model}), we
can generate an artificial subset that has similar rank-frequency
distribution as SFW. We start from a dataset that follows
the Zipf's law; then we go through the data points  from the top-rank ones 
down, and select rank-$r$ sample whenever its corresponding $r_{new}$,
according to Eq.(\ref{eq-cumsum1}), reaches a new integer value. 

Fig.\ref{fig2}(B) shows the rank-frequency plot for such selected
data points in log-log scale, with $N=15000$ words in the full set,
whose token counts follow a power-law with exponent $1.07$, and
20000 tokens for the top word. The BRF with $\alpha=0.93$ and $\beta=0.91$
fits the $T(r_{new})$ of the  artificially generated data well.

From Eq.(\ref{eq-r2rnew}), if we can derive a formula relating
the full-set rank $r$ and the subset rank $r_{new}$, we may have
an analytic expression for the rank-frequency relation in the subset,
as we know the full set follows a Zipf's law. In our subset selection
model (Eq.(\ref{eq-subset-model})), the $r_{new}$ is linked to $r$
through (the upper limit of) an integral.

In the limit of $r$ small (head area), $u/r_{mid}^\gamma << 1$,
and the integral leads to $r_{new} \approx \int^r du \sim r$. 
In the large $r$ limit (tail area), $u/r_{mid}^\gamma >> 1$,
the integral is:
\begin{equation}
r_{new} \approx r_{mid}^\gamma\int_1^r \frac{du}{u^\gamma}
= \frac{r_{mid}^\gamma}{-\gamma+1}  \left. \frac{1}{u^{\gamma-1}} \right\vert_1^r
= \frac{r_{mid}^\gamma}{\gamma-1} \left( 1-  \frac{1}{r^{\gamma-1}} \right) .
\end{equation}
Combining the two limits, we have this approximate relationship between
$r$ and $r_{new}$:
\begin{equation}
r = \left\{
\begin{array}{ll}
r_{new} & \mbox{head}\\
\sim \frac{1}{(R-r_{new})^{1/(\gamma-1)}} & \mbox{tail} \\
\end{array}
\right.
\end{equation}
where the coefficient $R$ is some combination of $r_{mid}$ and $\gamma$.
Inserting this relationship to the Zipf's law $1/r^\alpha$, we have
an approximation of the rank-frequency formula for the $r_{new}$:
\begin{equation}
 T(r_{new}) \sim \frac{ (R-r_{new})^{\alpha/(\gamma-1)}}{r_{new}^\alpha}
\end{equation}
which is a BRF with $\beta= \alpha/(\gamma-1)$.

\subsection{Rank-frequency plot of non-stopwords and non-function-words follow a quadratic function}

\begin{figure}[H]
 \begin{center}
 \includegraphics[width=1.0\textwidth]{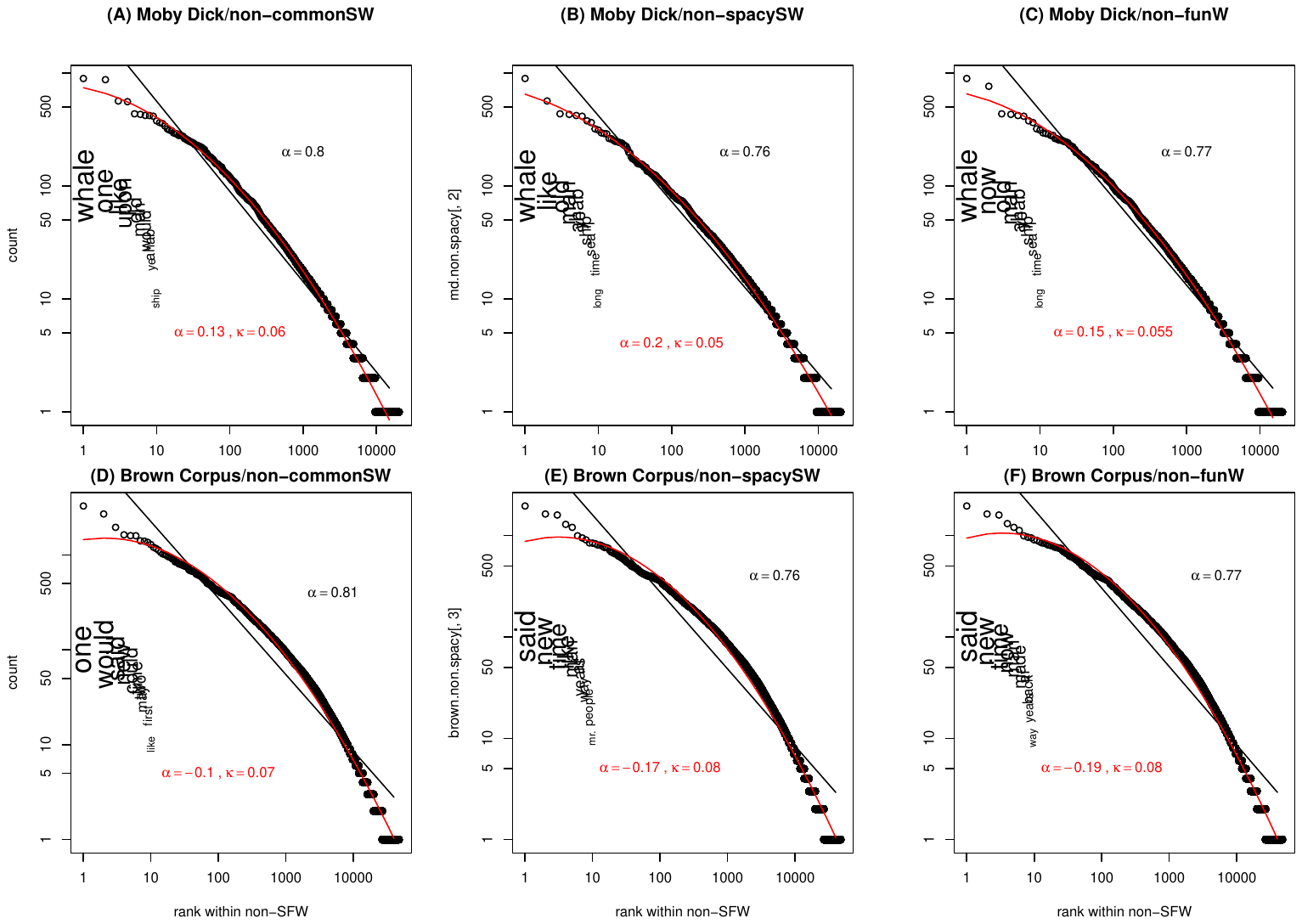}
 \end{center}
\caption{ \label{fig4}
Rank-frequency plots for non-stopwords and non-function-words. 
Top (bottom) row is for {\sl Moby Dick} ({\sl Brown corpus}),
and left (middle,right) column is for NLTK stopwords excluding contracted words
(spaCy stopwords, function words). 
The fitting function by power-law (Zipf's) function
is  in black, and those by quadratic function, 
$\log(T) \sim  - \alpha \log(r) - \kappa (\log(r))^2$
(Eq.(\ref{eq-quadratic})), are in red.  }
\end{figure}

\indent

In this subsection, we turn things around to examine the rank-frequency
plots of the remaining words (called ``non-stopwords" and ``non-function-words" or NSFW) 
after stopwords or function words being
removed. Fig.\ref{fig4} shows the six combinations
for three NSFW lists and for two sources of texts. It is clear that
none of them follows the inverse-power-law Zipf's law. The best attempts to fit
the data lead to inverse-power-law exponents of around 0.8, but the points
simply do not follow a straight line visually.

To fit the curved lines in Fig.\ref{fig4}, we have tested three nonlinear functions:
BRF (Eq.(\ref{eq-brf})), Mandelbrot function (Eq.(\ref{eq-mand})),
and quadratic equation (Eq.(\ref{eq-quadratic})). Table \ref{table1}
shows the adjusted $R^2$ for these three functions on the six
text/SFW combinations (note that these $R^2$s are calculated
on a geometrically sampled points from the NSFW).

\begin{table}[H]  
\begin{center}
Fitting performance of NSFW rank-frequency plots ($R^2$) \\
\begin{tabular}{c|c|c|c|c|c|c}
\hline
function & MD/common & MD/spaCy & MD/funW& Brown/common & Brown/spaCy &Brown/funW \\
\hline
Zipf 	& 0.967 & 0.973 
& 0.970 & 0.941  & 0.930 & 0.931 \\
BRF 	& 0.972 & 0.978 & 0.975 &0.953 & 0.944  & 0.944\\
Mandelbrot & 0.995 & 0.992 & 0.993 & 0.981  & 0.976  & 0.979 \\
quadratic & {\bf 0.998} & {\bf 0.997} & {\bf 0.998} & {\bf 0.994}  & {\bf 0.993} & {\bf 0.995}\\
\hline
\end{tabular}
\end{center}
\caption{ \label{table1}
Adjusted $R^2$ of four fitting functions of rank-frequency plot of NSFWs, on six 
combinations of text sources and SFW lists.
The four fitting functions are power-law (Zipf's law) Eq.(\ref{eq-zipf}),
Beta Rank Function (BRF) Eq.(\ref{eq-brf}),
Mandelbrot function  Eq.(\ref{eq-mand}), and
quadratic regression Eq.(\ref{eq-quadratic}).
The fitting is not carried out over all data points, but on a log-evenly-distributed
points. The best performing function is marked with bold.
}
\end{table}

It can be seen from Table \ref{table1} that quadratic function outperforms
other functions in all six situations. 
Although the coefficient for the  quadratic term $\kappa$ (0.05 $\sim$ 0.08) is
relatively smaller than that of the linear term $\alpha$ (0.13, 0.2
and 0.15 for Moby Dick, for power-law decrease, and $-0.1$, $-0.17$ 
and $-0.19$ for Brown Corpus, for power-law increase), it actually dominates the
shape of the fitting function. Even a change of sign in $\alpha$ from
Moby Dick to Brown Corpus does not greatly affect the shape of the
fitting line (except for the top ten data points).

BRF does not fit the NSFW data as it often does not bend until the very 
end of the tail (results not shown). The Mandelbrot function could be
a perfect choice if the top $B$ words are SFWs: 
it is because the top-ranking NSFWs would be the ($B+1$)-rank word in the full
set, the next NSFW would be the ($B+2$)-rank word in the full set, etc.
As the full set is assumed to follow the Zipf's law, the rank-frequency
plot for NSFWs, under this assumption, would be $1/(r_{new}+B)^\alpha$.
However, SFWs are more widely distributed and not just the top $B$ words,
making Mandelbrot function less ideal.

For Moby Dick text, the top ten NSFWs are 
``whale, one, 
like, upon, old, man, would, ahab, ye, ship", if the shared-in-common stopwords are
removed, and ``whale, like, old, man, ahab, ye, ship, sea, time, long" if
spaCy stopwords are removed, and ``whale, now, old, man, ahab, ye, ship,
sea, time, long" if the function words are removed.
These clearly are better keywords candidate for Moby Dick (e.g., whale, old, man, sea). 
Others include novel's character names (e.g. Ahab),
and ``ye" is an archaic word for ``you".

The Brown Corpus is a collection of many disciplines and genre, from
politics to technical works.  Therefore, the top NSFWs reflect
more on what is not in a particular stopword/function-word list. The top ten NSFWs
with the NLTK-spacy common stopwords removed are: 
``one, would, said, new, could, time,
two, may, first, like" and that with the spacy stopwords removed are:
``said, new, time, like, man, af, years, way, people, mr.",
and that with function words removed are:
``said, new, time, now, man, made, af, back, years, way".
The peculiar word of ``af" is actually Af, seemingly converted 
from $A_f$ in the technical documents.

\subsection{Further theoretical explorations of rank-frequency
plot of non-stopwords and non-function-words}

\begin{figure}[H]
 \begin{center}
 \includegraphics[width=1.0\textwidth]{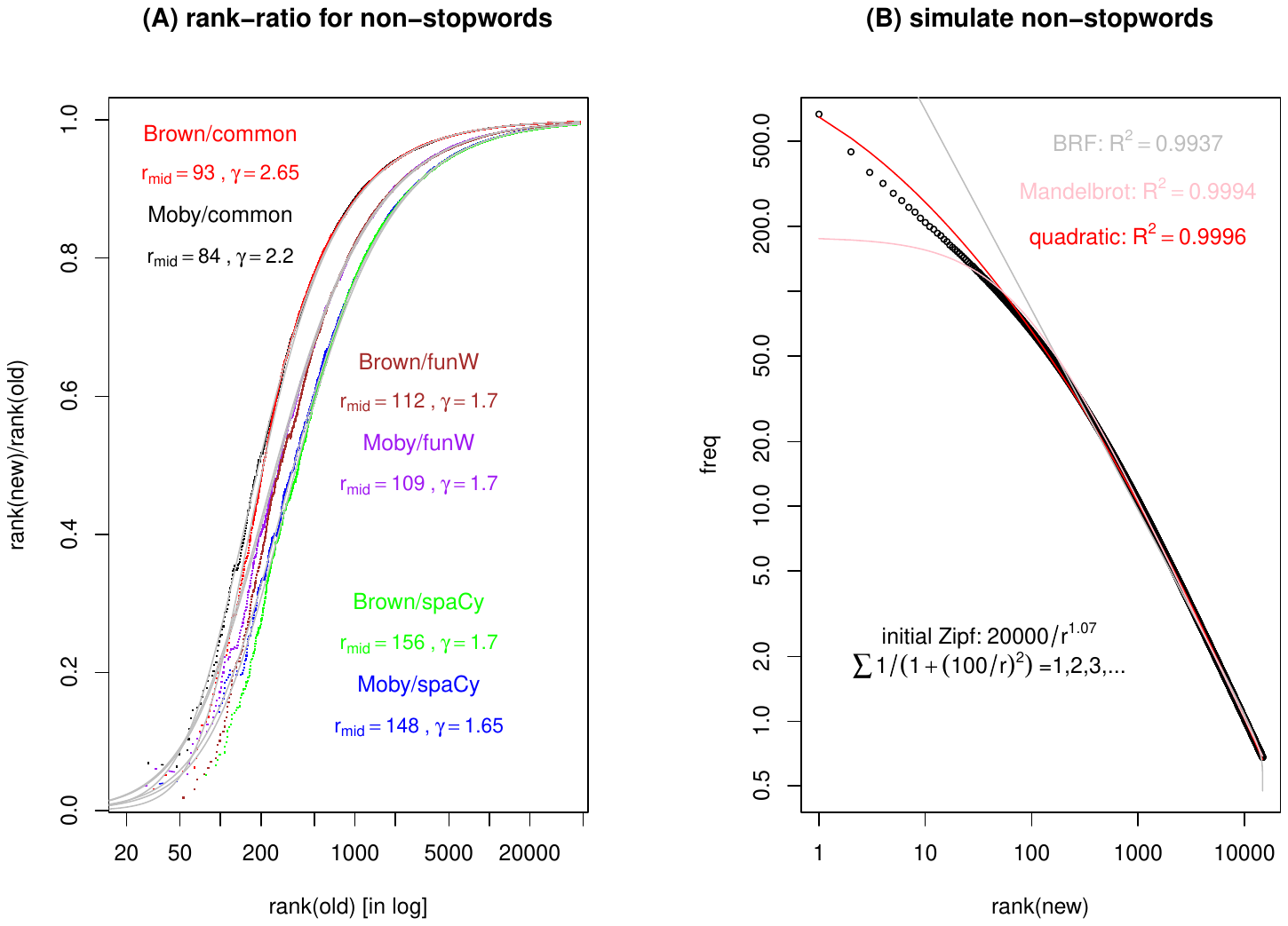}
 \end{center}
\caption{ \label{fig5}
The $r$ and ${r'}_{new}$ are the rank of
a word that does not belong to a stopword or a function word list 
(NSFW)  before  and after stopwords/function-words are removed.
(A) The ratio of these two ranks ${r'}_{new}/r$ for the six 
text source/SFW list
combinations. The fitting lines are the cumulative sum from
Eq.(\ref{eq-cumsum2}) with the best fitting parameters.
(B) Rank-frequency plot of an artificial dataset produced by picking
a word whenever its ${r'}_{new}$ according to Eq.(\ref{eq-cumsum2})
reaches a new integer. The three fitting lines are BRF (grey),
Mandelbrot function (pink), and quadratic function (red).
}
\end{figure}

\indent

From our subset selection model Eq.(\ref{eq-subset-model}),
the probability of a rank-$r$ not being selected is one minus the probability
to be selected:
\begin{equation}
\label{eq-not-model}
Prob(NSFW)_r = 1- \frac{1}{1+(r/r_{mid})^\gamma}
= \frac{(r/r_{mid})^\gamma}{1+(r/r_{mid})^\gamma}
= \frac{1}{1+(r_{mid}/r)^\gamma} .
\end{equation}
This is a Hill's equation (increasing or activating form of) \citep{hill}, 
where $\gamma$ is the Hill's coefficient. When $\gamma=1$,
the Hill's equation becomes Michaelis-Menten equation \citep{mm}.
The rank among non-stopwords $r'_{new}$ is then
the cumulative sum of the above probability: 
\begin{equation}
\label{eq-cumsum2}
r'_{new} = \sum_{u=1}^r \frac{1}{1+ ( \frac{r_{mid}}{u} )^\gamma}
\approx \int_{u=1}^{u=r} \frac{du}{1+ ( \frac{r_{mid}}{u})^\gamma}
\end{equation}
We use a brute force method to estimate the parameter values
for $r_{mid}$ and $\gamma$ for four different text/stopword combinations, 
which minimize the difference between  expected and observed 
$r'_{new}$ in log scale.  The resulting prediction lines fit the 
observed  $r'_{new}/r$ data very well (see Fig.\ref{fig5}(A)), with perhaps 
the exception of the head area for Brown corpus/spaCy stopword combination.

We can also simulate a NSFW dataset by choosing rank-$r$
samples whose $r'_{new}$ value by Eq.(\ref{eq-not-model}) reaches
an integer 1,2, 3, $\cdots$. Fig.\ref{fig5}(B) shows the rank-frequency
plot for such a simulated NSFW data from a full set that follows
the Zipf's law $1/r^{1.07}$. Again, the quadratic function provides
a better fit than both Mandelbrot function and BRF. 

Our subset model, or not-being-selected model, offers a hint on why 
the quadratic function may be a good fitting function.  At head region, 
$u << r_{mid}$, the integral in Eq.(\ref{eq-cumsum2}) becomes
in $r'_{new} \sim r^{\gamma+1}$ or $r \sim {r'_{new}}^{1/(\gamma+1)}$.
At the tail region, $u >> r_{mid}$, $r'_{new} \sim r$.
Plugging these to the Zipf's law $1/r^\alpha$ for the full set, 
we can see that the tail is a power-law $1/{r'_{new}}^\alpha$,
whereas the head is a power-law with a different exponent:
$1/{r'_{new}}^{1/(\gamma+1)}$. A quadratic function in log-log scale
would make a transition from one power-law to another possible.

\section{Discussions and conclusions}

\indent

{\bf Stopwords vs. function words:} In this paper, we encounter
two concepts, from two different fields, that are  nevertheless
related to each other. Stopwords originated from the NLP to operationally
describe ubiquitous words; and function words are certain categories of words
that play the role of grammatical glue. Both concepts can not lead
to an unambiguous list: (1) although ubiquity can be measured by
(the opposite of) inverse document frequency (e.g., \citep{miretie}),
it depends on which corpus is used, how a corpus is partitioned into
documents, as well as a cutoff threshold to the measure of ubiquity;
(2) for function words, some words can be either classified as  
function words in some situation, or not in other situations. 

However, there is an intrinsic reason why stopwords and function words
are closely related. A word can be ubiquitous (appear everywhere, in almost all 
documents) due to two explanations. One, that word plays a grammatical
role in constructing sentences and connecting words, i.e., it is
therefore  a function word. Two, that word describes a common concept, 
a common action, etc.  represented by a ubiquitous noun or verb, etc. 
That word is then a non-function-word-but-stopword. 

Of the 302 spaCy stopwords list, 218 of them are considered to
be function words and 84 are not. The ranking of these 218 function words
(in Brown corpus) are not significantly different from those of all spaCy stopwords
(Wilcoxon test $p$-value =0.018), whereas the ranking of the 84
non-function-word (but stopwords) are significantly different from
those of all spaCy stopwords (Wilcoxon test $p$-value = 1$\times 10^{-5}$).
The non-function-word-but-stopwords tend to have lower rankings.
This result indicates that the 218 function words within the spaCy stopwords
are representative of the whole list, validating our claim that
function words and stopwords are closely related. 

 
\indent

{\bf Short span on $x$-axis:}
Although we have shown that beta rank function is a perfect fitting
function for stopword's rank-frequency plot, it should be reminded
that the stopword lists are almost always short (100 $\sim 10^3$).
Fitting a short ranged rank-frequency data with functions of
one extra parameter is relatively easier. Previously we
found that many functions can fit the letter rank-frequency plot
well, where there are only 20$\sim$30 data points \citep{wli-letter}.
As a comparison, when the $x$-axis range expands, a previous
considered to be a good fitting function may end up to be not so good
after all.  For example, the Heaps' law (admittedly, not a rank-frequency
plot, but still a monotonically increasing plot) turns out to 
be not a good fitting function with the number of tokens 
($x$-axis) increase \citep{wli-heaps-dna,wli-heaps}.

\begin{figure}[H]
 \begin{center}
 \includegraphics[width=1.0\textwidth]{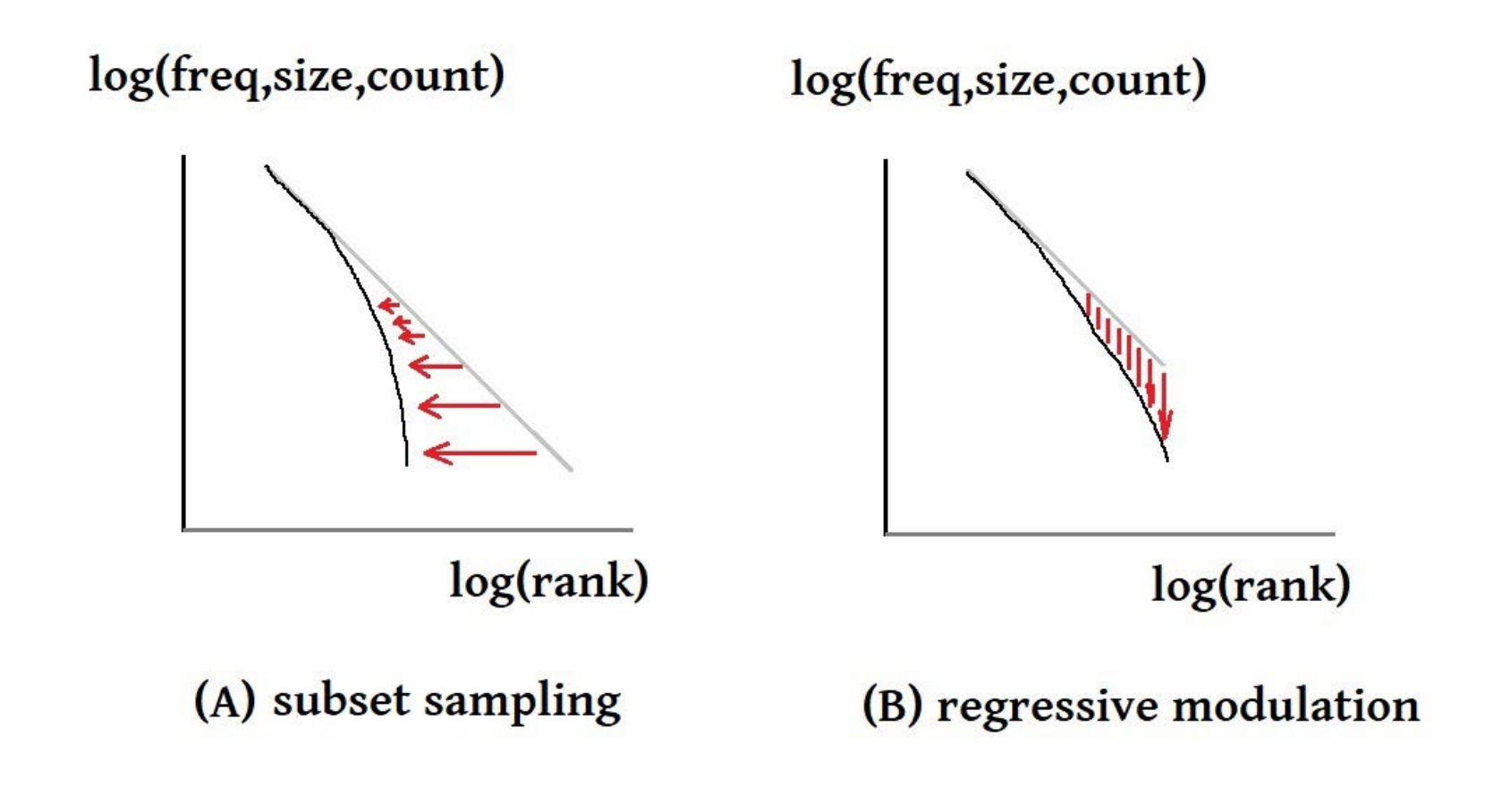}
 \end{center}
\caption{ \label{fig6}
Schematic illustration of two methods to produce a dataset
with BRF starting from a Zipf's law following dataset. 
The rank-frequency plot of the original dataset (grey line) is 
a straight line in log-log scale.
(A) The subset selection method discussed in Sec.\ref{sec-ssm}.
In this method, only a subset of the original dataset is selected.
(B) The regressive modulation method discussed in \citep{oscar26}.
In this method, a sample's value is reduced regressively
(the lower the value, the more reduction).
}
\end{figure}

{\bf Different ways to bend a straight line in log-log scale:}
From an almost perfect power-law (Zipf's) distribution of all words
to BRF distribution for stopwords, subset selection is a natural
mechanism  simply because stopwords are a subset of all words. 
Fig.\ref{fig6}(A) shows how a power-law function, which is a straight
line in log-log plot, may be converted to a BRF by subset sampling. 
Since the number of samples in the subset is lower than that in the
original dataset, the last point's rank ($x$-axis) is shrunk. On the other
hand, since top-ranking words are often remain in the subset, the head
of the rank-frequency plot remains at the straight line. With both ends
of the plot fixed, it must bend as a curve. Controlling the sampling 
rate by Eq.(\ref{eq-subset-model}) would sample less and less towards 
the tail of the distribution, which leads to BRF.

A completely different mechanism to produce data satisfying BRF
is the regressive modulation model illustrated in Fig.\ref{fig6}(B)
\citep{oscar26}. In this model, the initial rank-frequency distribution
is also power-law, but the number of samples remains the same.
Then each value (the $y$-axis) is reduced regressively, meaning,
the higher-ranking samples either remain the same or change very
little, whereas lower-ranking samples are modulated more.
In the economic policy context, regression tax system is the 
opposite of progressive one, and it taxes disproportionately
lower-income individuals.

Besides these two basic models for BRF in Fig.\ref{fig6}(A) and (B),
it is also possible to change both sample sizes and sample values.
In the split-merge model for gene sets \citep{wli16}, after a split
(merge), the number of gene sets increase (decreases); at the same
time, the size of gene sets decrease (increase), plus a re-ranking
of all gene sets.  In this case, a more sophisticated graph than Fig.\ref{fig6} is 
needed to describe the dynamic change from power-law towards BRF.

{\bf Implications to Zipf's law:} Our observation that non-stopwords
and non-function-words do not follow a strict Zipf's law (Fig.\ref{fig4})
means that if all function words or stopwords are removed from a 
written or spoken language, the words in the remaining language
no longer hold Zipf's law true.  Such lexical-centric language as versus
grammatical language does exist in unusual situations. Early childhood
language learners often skip function words or ignore different forms
of function words \citep{gerken}. A certain type of aphasia patients,
called agrammatism patients, are characterized by the omission of
grammatical or function words in their speech \citep{villiers,kean}.
All these can be called a ``telegraphic speech" \citep{goodglass,bloom}.

In a less extreme case, adults learning a second language (L2) go through
a stage with structural (as well as semantic) simplification \citep{ellis}.
In this stage, also called ``basic variety" in \citep{klein}, the language
beginners prioritize content words over function words: articles are
skipped, prepositions are rare, zero or limited verb morphemes are utilized.
These would lead to a situation where the token counts for function words
are much reduced, and Zipf's law may not hold.

Similar situations may appear when a data compression is necessary,
sometimes for cost consideration. For example, the telegraph text
itself \citep{shannon}, newspaper headlines \citep{koch}, etc.
As content words are usually more important than function words,
the latter is more likely to be skipped during a data compression.
Consequently, the frequencies of function words can be greatly
reduced, if not vanished, resulting in a non-Zipfian distribution.

{\bf Conclusions:}
In this work, we present a general picture on the
rank-frequency distribution of stopwords and/or function words.
While Zipf's law at the level of the complete vocabulary can be true,
when the vacabulary is partitioned into linguistically and computationally relevant
sub-vacabularies, Zipf's law is not expected to be preserved.
Ubiquitous, grammatically oriented, items are concentrated nonuniformly
along the global rank axis. A rank-dependent membership model,
a decreasing Hill's function, explains why their re-ranked frequencies
follow a Beta Rank Function. Similarly, the selection model for the
complementary lexical vocabulary, a direct Hill's function, lead to
a quadratic fitting function for the rank-frequency plot, which also deviates
from the pure power-law.
Using these results we predict that languages
that are deprived of function words, such as telegraphic speech,
do not follow Zipf's law for words.

\section*{Acknowledgment}
WT would like to thank Ying ZENG for discussions.

\newpage

\section*{Appendix: three stopword lists and one function words list}

\indent

The 122-stopword list which is the common set between NLTK and spaCy
stopword lists contains these words:
a, about, above, after, again, against, all, am, an, and, any, are, as, at, be, because, been, before, being, below, between, both, but, by, can, did, do, does, doing, down, during, each, few, for, from, further, had, has, have, he, her, here, hers, herself, him, himself, his, how, i, if, in, into, is, it, its, itself, just, me, more, most, my, myself, no, nor, not, now, of, off, on, once, only, or, other, our, ours, ourselves, out, over, own,  same, she, should, so, some, such, than, that, the, their, them, themselves, then, there, these, they, this, those, through, to, too, under, until, up, very, was, we, were, what, when, where, which, while, who, whom, why, will, with, you, your, yours, yourself, yourselves.

The 302-stopword list from spaCy contains these words (the 180 stopwords that are not in
the NLTK/spaCy common list are marked by bold font):  
a, about, above, {\bf across}, after, {\bf afterwards}, again, against, all, {\bf almost}, 
{\bf alone}, {\bf along}, {\bf already}, {\bf also, although, always}, am, {\bf among, amongst, amount}, 
an, and, {\bf another}, any, {\bf anyhow, anyone, anything, anyway, anywhere}, are, 
{\bf around}, as, at, {\bf back}, be, {\bf became}, because, {\bf become, becomes, becoming}, 
been, before, {\bf beforehand, behind}, being, below, {\bf beside, besides}, between, {\bf beyond}, 
both, {\bf bottom}, but, by, {\bf call}, can, {\bf  cannot,  could}, did, 
do, {\bf does}, doing, done, down, {\bf due}, during, each, {\bf eight, either, 
eleven, else, elsewhere, empty, enough, even, ever, every, everyone, everything, 
everywhere, except}, few, {\bf fifteen, fifty, first, five}, for, {\bf former, formerly, 
forty, four}, from, {\bf front, full}, further, {\bf get, give, go}, had, 
has, have, he, {\bf hence}, her, here, {\bf hereafter, hereby, herein, hereupon}, 
hers, herself, him, himself, his, how, {\bf however, hundred}, i, if, 
in, {\bf indeed}, into, is, it, its, itself, {\bf keep, last, latter, 
latterly, least, less}, just, {\bf made, make, many, may}, me, {\bf meanwhile, 
might, mine}, more, {\bf  moreover}, most, {\bf mostly, move, much, must}, my, 
myself, {\bf name, namely, neither, never, nevertheless, next, nine}, no, {\bf nobody, 
none, noone}, nor, not, {\bf nothing}, now, {\bf nowhere}, of, off, {\bf often}, 
on, once, {\bf one}, only, {\bf onto}, or, other, {\bf others, otherwise}, our, 
ours, ourselves, out, over, own, {\bf part, per, perhaps, please, put, 
quite, rather},  {\bf really, regarding}, same, {\bf say, see, seem, seemed, 
seeming, seems, serious, several}, she, should, {\bf show, side, since, six, 
sixty}, so, some, {\bf somehow, someone, something, sometime, sometimes, somewhere, still}, 
such, {\bf take, ten}, than, that, the, their, them, themselves, then, 
{\bf thence}, there, {\bf thereafter, thereby, therefore, therein, thereupon}, these, they, {\bf third}, 
this, those, {\bf though, three}, through, {\bf throughout,  thus}, to, {\bf together}, 
too, {\bf top, toward, towards, twelve, twenty, two}, under, until, up, 
{\bf unless, upon, us, used, using, various}, very, {\bf via}, was, we, 
{\bf well}, were, what, {\bf whatever}, when, {\bf whence, whenever}, where, {\bf whereafter, whereas, 
whereby, wherein, whereupon, wherever, whether}, which, while, {\bf whither}, who, {\bf whoever, 
whole}, whom, {\bf whose}, why, will, with, {\bf within, without, would, yet}, 
you, your, yours, yourself, yourselves.

The snowball stopword list with 175 words (all contracted words and three
words marked in bold are not in the above 123- and 305-stopword lists):
a, about, above, after, again, against, all, am, an, and, any, are, 
aren't, as, at, be, because, been, before, being, below, between, both, but, 
by, can't, cannot, could, couldn't, did, didn't, do, does, doesn't, doing, don't, 
down, during, each, few, for, from, further, had, hadn't, has, hasn't, have, 
haven't, {\bf having}, he, he'd, he'll, he's, her, here, here's, hers, herself, him, 
himself, his, how, how's, i, i'd, i'll, i'm, i've, if, in, into, 
is, isn't, it, it's, its, itself, let's, me, more, most, mustn't, my, 
myself, no, nor, not, of, off, on, once, only, or, other, {\bf ought}, 
our, ours, ourselves, out, over, own, same, shan't, she, she'd, she'll, she's, 
should, shouldn't, so, some, such, than, that, that's, the, their, {\bf theirs}, them, 
themselves, then, there, there's, these, they, they'd, they'll, they're, they've, this, 
those, through, to, too, under, until, up, very, was, wasn't, we, we'd, we'll, 
we're, we've, were, weren't, what, what's, when, when's, where, where's, which, while, 
who, who's, whom, why, why's, will, with, won't, would, wouldn't, you, you'd, 
you'll, you're, you've, your, yours, yourself, yourselves.  

The function words list we constructed is based on the spaCy and
NLTK stopword lists and a list from \citep{craig} with nouns, verbs, adjectives,
adverbs,etc. removed. It contains 228 entries):
a, about, above, across, after, afterwards, against, all, 
along, also, although, am, among, amongst, an, and, 
another, any, anyhow, anyone, anything, anyway, anywhere, are, 
as, at, be, because, been, before, behind, being, 
below, beside, besides, between, beyond, both, but, by, 
can, cannot, could, did, do, does, doing, done, 
down, during, each, either, else, enough, even, every, 
everyone, everything, except, few, first, five, for, four, 
from, had, has, have, having, he, hence, her, 
hereafter, hereupon, hers, herself, him, himself, his, how, 
however, i, if, in, into, is, it, its, itself, last, least, less, like, many, may, me, 
meanwhile, might, mine, more, moreover, most, much, must, my, myself, namely, neither, 
nevertheless, no, nobody, none, noone, nor, not, nothing, of, off, on, once, 
one, only, onto, or, other, others, otherwise, our, ours, ourselves, out, over, 
own, past, per, quite, rather, regarding, round, same, several, shall, she, should, 
since, six, so, some, someone, something, somewhat, still, such, ten, than, that, 
the, their, theirs, them, themselves, then, there, thereafter, thereby, therefore, thereupon, these, 
they, third, this, those, though, three, through, throughout, thus, till, to, too, 
toward, towards, two, under, unless, until, unto, up, upon, us, very, via, 
was, we, were, what, whatever, when, whence, whenever, where, whereafter, whereas, whereby, 
wherein, whereupon, wherever, whether, which, while, whilst, whither, who, whoever, whom, whose, 
why, will, with, within, without, would, yet, you, your, yours, yourself, yourselves.

\end{document}